\documentclass[conference]{IEEEtran}
\IEEEoverridecommandlockouts
\usepackage{circuitikz}
\usepackage{graphicx}
\usepackage{comment}
\usepackage{subfig}
\usepackage[normalem]{ulem} % underline without breaking
\usepackage{amstext}
\usepackage{pifont}
\usepackage{MathDefs}
\usepackage{docmute}
\usepackage{amsmath,amsfonts,amssymb,amsthm}
\usepackage{hyperref}
\usepackage{cite}
\usepackage{balance}

\hypersetup{
	colorlinks  = true, %Colours links instead of ugly boxes
	urlcolor     = blue, %Colour for external hyperlinks
	linkcolor   = blue, %Colour of internal links
	citecolor   = red %Colour of citations
}
\usepackage{nomencl}
\makenomenclature
\usepackage{color, soul} % for color text

\newcommand{\blue}{ \textcolor{blue} }

\usepackage{tcolorbox} % for box arund eqn and text
\newtcolorbox{Educ}[1]{
	title=#1,
	beamer, 
	colback=xlightblue,
	colframe=blue!30,
	fonttitle=\bfseries,
	left=1mm,
	right=1mm,
	top=1mm,
	bottom=1mm,
	middle=1mm,
	breakable,
}

\usepackage{url}
\usetikzlibrary{positioning}
\usetikzlibrary{arrows.meta, arrows}

\usepackage{algorithm}
\usepackage{algpseudocode} % with end
\usepackage{matlab-prettifier}

\usepackage{tabularx}
\newcolumntype{L}[1]{>{\raggedright\let\newline\\\arraybackslash\hspace{0pt}}m{#1}}
\newcolumntype{C}[1]{>{\centering\let\newline\\\arraybackslash\hspace{0pt}}m{#1}}
\newcolumntype{R}[1]{>{\raggedleft\let\newline\\\arraybackslash\hspace{0pt}}m{#1}}

\usepackage{pgfplots}
\DeclareMathOperator{\Tr}{Tr}

\newcommand\copyrighttext{%
	\footnotesize \textcopyright 2026 IEEE. Personal use of this material is permitted.  Permission from IEEE must be obtained for all other uses, in any current or future media, including reprinting/republishing this material for advertising or promotional purposes, creating new collective works, for resale or redistribution to servers or lists, or reuse of any copyrighted component of this work in other works.
	DOI: \href{https://ieeexplore.ieee.org/xpl/conhome/1000362/all-proceedings}{pending}}
\newcommand\copyrightnotice{%
	\begin{tikzpicture}[remember picture,overlay]
		\node[anchor=south,yshift=10pt] at (current page.south) {\fbox{\parbox{\dimexpr\textwidth-\fboxsep-\fboxrule\relax}{\copyrighttext}}};
	\end{tikzpicture}%
}

\begin{document}

\title{An Objective Performance Evaluation of the LSTM Networks in Time Series Classification}

\author{\IEEEauthorblockN{1\textsuperscript{st} Sooraj Sunil}
	\IEEEauthorblockA{\textit{Electrical and Computer Engineering} \\
		\textit{University of Windsor}\\
		Windsor, ON, Canada \\
		sunil11@uwindsor.ca}
	\and
	\IEEEauthorblockN{2\textsuperscript{nd} Balakumar Balasingam}
	\IEEEauthorblockA{\textit{Electrical and Computer Engineering} \\
		\textit{University of Windsor}\\
		Windsor ON, Canada\\
		singam@uwindsor.ca}
	
	\thanks{
		The work of Balakumar Balasingam was supported by the Natural
		Sciences and Engineering Research Council of Canada (NSERC) under Grant
		RGPIN-2024-04557.
	}
}

\maketitle

\copyrightnotice
\begin{abstract}
The rapid adoption of deep learning has increasingly led to 
data-driven models replacing classical model-based algorithms, 
even in domains governed by well-understood physical laws.
While data-driven models, such as long short-term memory (LSTM) 
networks, have become a popular choice for time-series analysis, 
their performance relative to model-based approaches in 
structured environments is rarely evaluated objectively. This 
paper presents a performance evaluation framework comparing an 
LSTM classifier against a model-based expectation-maximization 
(EM) classifier for binary time-series classification. The 
evaluation is conducted on two scalar linear Gaussian 
state-space models differing only in their noise statistics, 
where the Kalman filter likelihood ratio test with true 
parameters serves as a reference for the best achievable 
classification performance. Through Monte Carlo simulations, 
the classifiers are evaluated across three axes: task 
difficulty, controlled by the separation in process or 
measurement noise between the two models; sequence length; 
and training dataset size. The results show that the EM 
classifier, which exploits the known model structure, performs 
strongly when the data conform to the assumed model class. The 
LSTM classifier requires a larger separation in noise statistics 
to achieve reliable classification, and its performance 
saturates below the reference classifier when the models differ 
only in measurement noise, regardless of sequence length or 
training dataset size. 
\end{abstract}


%\tableofcontents

\section{Introduction}
Classification of time-series data is a foundational problem 
across many domains, with applications in transportation 
systems~\cite{gupta2020early}, fault 
diagnosis~\cite{he2025multichannel}, data 
mining~\cite{ismail2019deep}, smart 
manufacturing~\cite{farahani2025time}, and biomedical 
engineering~\cite{wu2026mscgn}. In many of these settings, 
the underlying process is not directly observable and must be 
inferred from noisy measurements.

When the system dynamics are known and linear Gaussian, the 
Kalman filter provides the mathematically optimal framework for 
state estimation~\cite{kalman1960new}. However, in practice, 
the true signal model and noise statistics are rarely known 
exactly. Using incorrect model or noise statistics can lead to 
large estimation errors or even filter 
divergence~\cite{bar2004estimation}. Estimation under such 
uncertainty is termed \emph{adaptive estimation}, and is 
generally carried out in a suboptimal 
fashion~\cite{mehra2003approaches}. Among the many approaches, 
the expectation-maximization (EM) algorithm jointly estimates 
the unknown parameters and latent state sequence by iteratively 
maximizing the expected log-likelihood of the 
observations~\cite{Dempster1977,shumway1982approach}.

In recent years, deep learning methods have seen rapid adoption 
for time-series analysis~\cite{ismail2019deep,farahani2025time}. 
While long short-term memory (LSTM) networks have demonstrated 
strong performance in time-series 
forecasting~\cite{siami2018comparison,siami2019performance,
	essien2020deep}, convolutional neural networks (CNNs) and 
hybrid CNN-LSTM architectures have emerged as the preferred 
choice for time-series 
classification~\cite{karim2017lstm,karim2019multivariate}. 
These models capture temporal dependencies directly from data, 
without requiring explicit model assumptions. However, they 
operate as black-box models with limited interpretability, and 
their performance is heavily dependent on the availability of 
large labeled training datasets~\cite{kabkab2016size}. To 
address these limitations, hybrid model-based deep learning 
methods have begun to emerge, combining the structural knowledge 
of physical models with the flexibility of data-driven 
approaches~\cite{forti2023model}.

In this paper, we propose an objective performance evaluation 
framework for deep learning models applied to time-series 
classification. Specifically, we evaluate the LSTM classifier 
against the EM algorithm in a binary classification problem, 
where the goal is to distinguish between two univariate linear 
Gaussian state-space models sharing the same structure but 
differing in their noise statistics. Since the true model is 
known, the Kalman filter likelihood ratio test (LRT) with true 
parameters serves as a reference for the best achievable 
classification performance. The deviation of each classifier 
from this reference is quantified as a function of task 
difficulty, training set size, and sequence length. Unlike 
existing studies that benchmark deep learning models against 
classical methods such as the autoregressive integrated moving 
average (ARIMA) model~\cite{siami2018comparison,
	siami2019performance} on real-world datasets — where the 
optimal classifier is unknown — we evaluate on synthetically 
generated data from a known model, enabling an absolute 
performance comparison. While we focus on LSTM, the proposed 
framework is general and can be extended to any deep learning 
model for time-series classification or forecasting.

The remainder of this paper is organised as follows. 
Section~\ref{sec:Problem} defines the problem formulation. 
Section~\ref{sec:em} presents the model-based EM classifier. 
Section~\ref{sec:lstm} details the LSTM architecture. 
Section~\ref{sec:Results} presents the simulation results, 
followed by conclusions in Section~\ref{sec:conclusion}.

\section{Problem Definition}
\label{sec:Problem}
The system considered in this paper is a discrete-time (indexed by 
$k$) linear time-invariant (LTI) dynamic system described by the 
following state-space model~\cite{bar2004estimation}:
\begin{equation} 
		\label{eq:gen_ssm}
	\begin{aligned}
		\bx_{k} &= \bF \bx_{k-1} + \bv_k, \\
		\bz_k   &= \bH \bx_k + \bw_k,
	\end{aligned}
\end{equation}
where $\bx_k \in \mathbb{R}^{m_x}$ denotes the state vector, $\bF 
\in \mathbb{R}^{m_x \times m_x}$ is the state transition matrix, 
$\bz_k \in \mathbb{R}^{m_z}$ denotes the observation vector, $\bH 
\in \mathbb{R}^{m_z \times m_x}$ is the observation matrix, and 
$\bv_k$ and $\bw_k$ represent the process and measurement noises, 
respectively. The noise sequences are assumed to be zero-mean, white, 
and mutually independent Gaussian processes:
\begin{equation}
	\label{eq:covariance}
	\begin{aligned}
		\bv_k &\sim \mathcal{N}(\mathbf{0}, \bQ), \\
		\bw_k &\sim \mathcal{N}(\mathbf{0}, \bR),
	\end{aligned}
\end{equation}
where $\bQ \succeq 0$ and $\bR \succ 0$ denote the process and 
measurement noise covariance matrices. The initial state $\bx_0$ is 
assumed Gaussian with known mean $\bmu_0$  and covariance  $\bSigma_0$, and independent of 
the noise sequences. Two candidate models are considered:
\begin{itemize}
	\item {Model 1:} $\bF = \bF_1$, $\bH = \bH_1$, 
	$\bQ = \bQ_1$, $\bR = \bR_1$
	\item {Model 2:} $\bF = \bF_2$, $\bH = \bH_2$, 
	$\bQ = \bQ_2$, $\bR = \bR_2$
\end{itemize}
Each model induces a distinct distribution over observation sequences 
$\{\bz_1, \bz_2, \ldots, \bz_T\}$.

\subsection{Binary Classification Objective}
\label{subsec:binary_classify}
Let $\cZ_T = \{\bz_1, \bz_2, \ldots, \bz_T\}$ denote a finite-length observation sequence of length $T$. Given $\cZ_T$ generated by one of the two candidate models, the objective is to determine which model most likely produced the data. 
Formally, this corresponds to a binary hypothesis testing problem:
\begin{align}
	\begin{aligned}
			\mathcal{H}_1 &:\cZ_T \sim p(\cZ_T \mid \bF_1, \bH_1, \bQ_1, \bR_1),  \\
		\mathcal{H}_2 &: \cZ_T \sim p(\cZ_T\mid \bF_2, \bH_2, \bQ_2, \bR_2).
	\end{aligned}
\end{align}

The optimal Bayesian decision rule (under equal priors and 0--1 loss) 
is the LRT, which selects the model with the larger log-likelihood:
\begin{equation}
	\hat{i} = \arg\max_{i \in \{1,2\}} \log\, p(\cZ_T\mid 
	\bF_i, \bH_i, \bQ_i, \bR_i).
	\label{eq:lrt}
\end{equation}
For a linear Gaussian state-space model, the log-likelihood in the LRT is evaluated 
analytically using the Kalman filter innovation sequence:
\begin{equation}
	\log\, p(\cZ_T) = -\frac{1}{2}\sum_{k=1}^{T} 
	\left[ \log(2\pi S_k) + \frac{\upsilon^2}{S_k} \right],
	\label{eq:logLRT_eval}
\end{equation}
where $\upsilon_k$ is the innovation 
and $S_k$ is the innovation 
covariance, both computed during the Kalman filter forward pass.

In practice, the model parameters and noise statistics are unknown 
and must be estimated from labeled training data. This paper 
investigates two fundamentally distinct approaches:
\begin{enumerate}
	\item \textbf{Model-based (EM + Kalman LRT):} The parameters of 
	each candidate state-space model are estimated from $N_{\rm train}$ training 
	sequences $\cZ_T$ using the EM algorithm. Classification is 
	then performed via the Kalman filter LRT~\eqref{eq:lrt} using 
	the estimated parameters.
	
	\item \textbf{Data-driven (LSTM):} An LSTM network is trained on 
	$N_{\rm train}$ labeled sequences $\cZ_T$ to map observations 
	directly to model labels $i \in \{1,2\}$, without any assumptions 
	about the underlying model structure.
\end{enumerate}

The goal of this work is to compare these two approaches under 
controlled simulation conditions where the ground-truth parameters 
are known. Under these conditions, the LRT~\eqref{eq:lrt} with true 
parameters serves as an upper bound on classification accuracy, 
allowing the performance gap between each classifier and the 
theoretical optimum to be precisely quantified.

\section{The EM Algorithm}
\label{sec:em}

This section reviews the EM algorithm for estimating the parameters 
of the state-space model in~\eqref{eq:gen_ssm}, following the standard development 
in~\cite{Dempster1977,shumway1982approach,shumway2011time}.

 The basic idea is that 
  if the state sequence 
 $\cX_{T+1}=\{\bx_k\}_{k=0}^T$ were available alongside the 
 observations $\cZ_T=\{\bz_k\}_{k=1}^T$, the complete data 
 $\{\cX_{T+1}, \cZ_T\}$ would have joint density
\begin{equation}
	p\left(\cZ_T, \cX_{T+1} | \bTheta \right) = p(\bx_0) 
	\prod_{k=1}^T p(\bx_k | \bx_{k-1}) 
	\prod_{k=1}^T p(\bz_k|\bx_k),
	\label{eq:pdf}
\end{equation}
where $\bTheta = \{ \bQ, \bR, \bF, \bH, \bmu_0, \bSigma_0 \}$ 
denotes the full parameter set. Under the Gaussian assumptions and 
ignoring constants, the complete-data negative log-likelihood 
is~\cite{shumway2011time}:
\begin{equation}
	\begin{aligned}
		-2\ln p (\cZ_T, &\cX_{T+1}| \bTheta) = \ln |{\bSigma}_0|  + T \ln |\bQ| + T \ln |\bR| \\ 
		& + (\bx_0- \bmu_0)^\top 
		\bSigma_0^{-1}(\bx_0-\bmu_0)  \\
		&+ \sum_{k=1}^{T} (\bx_k- \bF\bx_{k-1})^\top 
		\bQ^{-1}(\bx_k - \bF\bx_{k-1})  \\
		&+ \sum_{k=1}^{T} (\bz_k- \bH\bx_k)^\top 
		\bR^{-1} (\bz_k- \bH\bx_k). 
	\end{aligned}
	\label{eq:logliklihoodGeneral}
\end{equation}

If the complete data were observed, the maximum likelihood estimates 
(MLEs) of $\bTheta$ could be obtained directly from multivariate 
normal theory~\cite{bar2004estimation,shumway2011time}. Since only $\cZ_T$ is available, the EM algorithm 
provides an iterative procedure for computing the MLEs by 
successively maximizing the conditional expectation of the 
complete-data log-likelihood. At iteration $i$, this expectation is:

\begin{equation}
	\cQ(\bTheta | \bTheta^{(i-1)}) = E\left\{ \log p(\cZ_T, 
	\cX_{T+1}| \bTheta) \big|  \cZ_T, \bTheta^{(i-1)} \right\}.
	\label{eq:Q_EM}
\end{equation}

\subsection*{Expectation Step}
The \emph{expectation step} computes the expected complete-data log-likelihood under 
the current parameter estimates $\bTheta^{(i-1)}$.
Since the state 
sequence $\cX_{T+1}$ is unobserved, the expectation is taken with 
respect to the conditional distribution $p(\cX_{T+1} \mid \cZ_T, 
\bTheta^{(i-1)})$, which is computed via the Kalman filter and the smoother~\cite{bar2004estimation,shumway2011time}. 
Specifically, the Kalman filter runs forward in time to compute the 
filtered estimates $\bx_{k|k}$ and $\bP_{k|k}$, and the  smoother 
runs backward to compute the smoothed estimates $\bx_{k|T}$, 
$\bP_{k|T}$, and the cross-covariance $\bP_{k,k-1|T}$, all under 
$\bTheta^{(i-1)}$;  for brevity, this 
dependence is not explicitly displayed.
Evaluating~\eqref{eq:Q_EM} yields:
{\small
\begin{equation}
	\begin{aligned}
		&\cQ(\bTheta | \bTheta^{(i-1)}) 
		= \ln |{\bSigma}_0| + \ln |{\bR}| + \ln |\bQ| \\ 
		&+ \Tr \left\{ \bSigma_0^{-1} \left[ \bP_{0|T} + 
		(\bx_0- \bmu_0)(\bx_0-\bmu_0)^\top \right] \right\} \\
		&+ \Tr \left\{ \bQ^{-1} \left[ \bS_{11} - \bS_{10}\bF^\top 
		- \bF\bS^{\top}_{10} + \bF\bS_{00} \bF^\top 
		\right] \right\} \\  
		&+ \Tr \left\{ \bR^{-1} \left[ \bM_{11} - \bM_{10}\bH^\top 
		- \bH\bM^{\top}_{10} + \bH\bM_{00} \bH^\top 
		\right] \right\},
	\end{aligned}
	\label{eq:Estep}
\end{equation}
}
where
\begin{align}
	\bS_{11} &= \sum_{k=1}^{T} \left( \bx_{k|T} \bx_{k|T}^\top 
	+ \bP_{k|T} \right), \label{eq:S11}\\
	\bS_{10} &= \sum_{k=1}^{T} \left( \bx_{k|T} \bx_{k-1|T}^\top 
	+ \bP_{k,k-1|T} \right), \\
	\bS_{00} &= \sum_{k=1}^{T} \left( \bx_{k-1|T}\bx_{k-1|T}^\top 
	+ \bP_{k-1|T} \right), \\
	\bM_{11} &= \sum_{k=1}^{T} \bz_k \bz_k^\top, \\
	\bM_{10} &= \sum_{k=1}^{T} \bz_k \bx_{k|T}^\top, \\
	\bM_{00} &= \sum_{k=1}^{T} \left( \bx_{k|T}\bx_{k|T}^\top 
	+ \bP_{k|T} \right).
\end{align}
Note that the smoothed quantities $\bx_{k|T}$, $\bP_{k|T}$, and 
$\bP_{k,k-1|T}$ are evaluated under the current parameter estimates 
$\bTheta^{(i-1)}$; for brevity, this dependence is not explicitly 
displayed in the expressions above

\subsection*{Maximization Step}
The \emph{maximization step} maximizes $\cQ(\bTheta \mid \bTheta^{(i-1)})$ with 
respect to each parameter in $\bTheta$. 
Since~\eqref{eq:Estep} 
decouples across the parameter blocks $\{\bF, \bQ\}$, $\{\bH, 
\bR\}$, and $\{\bmu_0, \bSigma_0\}$, each block can be maximized 
independently. Setting the derivatives to zero yields closed-form parameter
updates:
\begin{align}
	\widehat{\bF}^{(i)} &= \bS_{10}\bS_{00}^{-1}, 
	\label{eq:maxFGen} \\
	\widehat{\bQ}^{(i)} &= \frac{1}{T} \left[ \bS_{11} 
	- \bS_{10}\bS_{00}^{-1}\bS_{10}^\top \right], \\
	\widehat{\bH}^{(i)} &= \bM_{10}\bM_{00}^{-1}, 
	\label{eq:maxHGen} \\
	\widehat{\bR}^{(i)} &= \frac{1}{T} \left[ \bM_{11} 
	- \bM_{10}\bM_{00}^{-1}\bM_{10}^\top \right],
\end{align}
with the initial state and covariance updated as 
$\widehat{\bmu}_0^{(i)} = \bx_{0|T}$ and 
$\widehat{\bSigma}_0^{(i)} = \bP_{0|T}$. The expectation and maximization steps are 
iterated until the change in the estimated parameters falls below a 
prescribed threshold, indicating convergence.

\subsection{EM Assisted Classification of Time Series Data}
\label{sec:model_based_class}

For the binary classification problem in 
Subsection~\ref{subsec:binary_classify}, EM is run separately on the labeled 
training sequences of each class, producing two parameter estimates:
\begin{align}
	\widehat{\bTheta}_1 &= \{\widehat{\bF}_1, \widehat{\bH}_1, 
	\widehat{\bQ}_1, \widehat{\bR}_1, \widehat{\bmu}_{0,1}, 
	\widehat{\bSigma}_{0,1}\}, \\
	\widehat{\bTheta}_2 &= \{\widehat{\bF}_2, \widehat{\bH}_2, 
	\widehat{\bQ}_2, \widehat{\bR}_2, \widehat{\bmu}_{0,2}, 
	\widehat{\bSigma}_{0,2}\}.
\end{align}
A test sequence $\cZ_T$ is then classified via the LRT as:
\begin{equation}
	\hat{i} = \arg\max_{i\in\{1,2\}} \log p(\cZ_T \mid 
	\widehat{\bTheta}_i),
	\label{eq:MBdecision}
\end{equation}
where the log-likelihood is evaluated using the Kalman filter 
innovation sequence in~\eqref{eq:logLRT_eval}.

\section{The LSTM Network}
\label{sec:lstm}
LSTM networks are a class of recurrent neural networks designed to 
learn temporal dependencies in sequential data. Unlike standard 
recurrent networks, LSTM networks employ a gating mechanism that 
mitigates the vanishing and exploding gradient 
problems~\cite{hochreiter2001gradient}, enabling effective training 
over long sequences. A comprehensive review of LSTM variants can be 
found in~\cite{greff2016lstm}.

\begin{figure}[h]
	\centering
	\begin{tikzpicture}[
		block/.style={
			draw, 
			rectangle, 
			minimum width=3cm,  % This acts as the vertical height due to rotation
			minimum height=1.2cm, % This acts as the horizontal width due to rotation
			align=center, 
			font=\small,
			fill=gray!10,
			rounded corners=2pt,
			rotate=90             % Rotating the node itself allows native line breaks
		},
		arrow/.style={->, >=stealth, thick}
		]
		
		% Nodes with two-line text
		\node[block] (input) at (0,0)   {Sequence Input\\Layer};
		\node[block] (lstm)  at (2.0,0) {LSTM Layer};
		\node[block] (fc)    at (4.0,0) {Fully Connected\\Layer};
		\node[block] (soft)  at (6.0,0) {Softmax Layer};
		
		% Arrows
		\draw[arrow] (-1.2,0) -- (input);
		\draw[arrow] (input)  -- (lstm);
		\draw[arrow] (lstm)   -- (fc);
		\draw[arrow] (fc)     -- (soft);
		\draw[arrow] (soft)   -- (7.2,0);
	\end{tikzpicture}
	\caption{LSTM network architecture for sequence-to-label classification.}
	\label{fig:lstm_arch}
\end{figure}
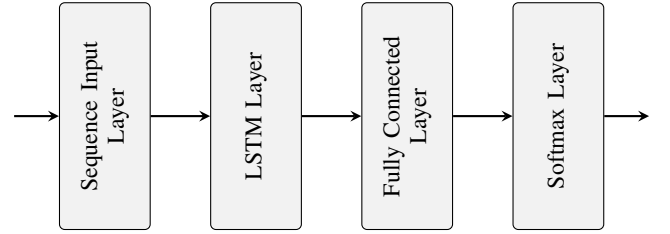

In this paper, we deploy an LSTM network for sequence-to-label 
classification. As illustrated in Fig.~\ref{fig:lstm_arch}, the 
network consists of a sequence input layer, an LSTM layer, a fully 
connected layer, and a softmax output layer.

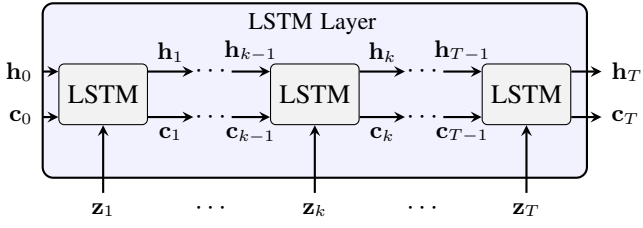
\begin{figure}[h]
	\centering
	\begin{tikzpicture}[
		block/.style={
			draw, rectangle, rounded corners=2pt,
			minimum width=1.0cm, minimum height=0.8cm,
			align=center, fill=gray!10
		},
		arr/.style={->, >=stealth, thick},
		]
		
		% Big rectangle LSTM layer 
		\draw[draw=black, thick, rounded corners=4pt, fill=blue!5]
		(-0.8, -1.1) rectangle (6.4, 1.2);
		\node[font=\small, anchor=north west] at (1.8, 1.2) {LSTM Layer};
		
		% LSTM cells
		\node[block] (lstm1)  at (0,   0) {LSTM};
		\node[block] (lstmK)  at (2.8, 0) {LSTM};
		\node[block] (lstmT)  at (5.6, 0) {LSTM};
		
		% Dots between lstm1 and lstmK
		\node at (1.45,  0.3) {$\cdots$};
		\node at (1.45, -0.3) {$\cdots$};
			\node at (1.45, -1.5) {$\cdots$};
	
		% Dots between lstmK and lstmT
		\node at (4.25,  0.3) {$\cdots$};
		\node at (4.25, -0.3) {$\cdots$};
		\node at (4.25, -1.5) {$\cdots$};

		% Inputs z_k
		\draw[arr] (0,   -1.3) node[below, font=\small]{$\bz_1$} -- (lstm1);
		\draw[arr] (2.8, -1.3) node[below, font=\small]{$\bz_k$} -- (lstmK);
		\draw[arr] (5.6, -1.3) node[below, font=\small]{$\bz_T$} -- (lstmT);
		
		% Hidden state h (top track)
		\draw[arr] (-0.8, 0.3) node[left, font=\small]{$\bh_0$}
		-- (lstm1.west  |- 0, 0.3);
		\draw[arr] (lstm1.east  |- 0, 0.3)
		-- ++(0.6, 0) node[midway, above, font=\small]{$\bh_1$};
		\draw[arr] (1.7, 0.3)
		-- (lstmK.west  |- 0, 0.3) node[midway, above, font=\small]{$\bh_{k-1}$};
		\draw[arr] (lstmK.east  |- 0, 0.3)
		-- ++(0.6, 0) node[midway, above, font=\small]{$\bh_{k}$};
		\draw[arr] (4.5, 0.3)
		-- (lstmT.west  |- 0, 0.3) node[midway, above, font=\small]{$\bh_{T-1}$};
		\draw[arr] (lstmT.east  |- 0, 0.3)
		-- ++(0.4, 0) node[right, font=\small]{$\bh_T$};
		
		% Cell state c (bottom track)
		\draw[arr] (-0.8, -0.3) node[left, font=\small]{$\bc_0$}
		-- (lstm1.west  |- 0,-0.3);
		\draw[arr] (lstm1.east  |- 0,-0.3)
		-- ++(0.6, 0) node[midway, below, font=\small]{$\bc_1$};
		\draw[arr] (1.7, -0.3)
		-- (lstmK.west  |- 0,-0.3) node[midway, below, font=\small]{$\bc_{k-1}$};
		\draw[arr] (lstmK.east  |- 0,-0.3)
		-- ++(0.6, 0) node[midway, below, font=\small]{$\bc_{k}$};
		\draw[arr] (4.5, -0.3)
		-- (lstmT.west  |- 0,-0.3) node[midway, below, font=\small]{$\bc_{T-1}$};
		\draw[arr] (lstmT.east  |- 0,-0.3)
		-- ++(0.4, 0) node[right, font=\small]{$\bc_T$};
		
	\end{tikzpicture}
	\caption{Data flow through an LSTM layer.}
	\label{fig:lstm_unrolled}
\end{figure}

Given an observation sequence $\cZ_T = \{\bz_1, \bz_2, \ldots, \bz_T\}$ 
of length $T$, the sequence input layer feeds the data into the network 
and applies data normalization. The normalized sequence is then passed 
to the LSTM layer as shown in Fig.~\ref{fig:lstm_unrolled}, where at 
each time step $k$, the LSTM cell updates its hidden state $\bh_k$ and 
cell state $\bc_k$ via the following gating mechanisms:
\begin{equation}
	\begin{aligned}
		\mathbf{i}_k &= \sigma_{\rm g}(\bW_{\rm i} \bz_k + \bR_{\rm i} \bh_{k-1} + \bb_{\rm i}), \\
		\mathbf{f}_k &= \sigma_{\rm g}(\bW_{\rm f} \bz_k + \bR_{\rm f} \bh_{k-1} + \bb_{\rm f}), \\
		\mathbf{o}_k &= \sigma_{\rm g}(\bW_{\rm o} \bz_k + \bR_{\rm o} \bh_{k-1} + \bb_{\rm o}), \\
		\tilde{\bc}_k &= \tanh(\bW_{\rm c}\bz_k + \bR_{\rm c} \bh_{k-1} + \bb_{\rm c}), \\
		\bc_k &= \mathbf{f}_k \odot \bc_{k-1} + \mathbf{i}_k \odot \tilde{\bc}_k, \\
		\bh_k &= \mathbf{o}_k \odot \tanh(\bc_k),
	\end{aligned}
\end{equation}
where $\mathbf{i}_k$, $\mathbf{f}_k$, and $\mathbf{o}_k$ denote the input, 
forget, and output gates, respectively; $\tilde{\bc}_k$ is the candidate 
cell state; $\sigma_{\rm g}(\cdot)$ is the sigmoid gate activation function; 
$\tanh(\cdot)$ is the hyperbolic tangent state activation function; and 
$\odot$ denotes the element-wise (Hadamard) product.

The final hidden state $\bh_T \in \mathbb{R}^{n_h}$, where $n_h$ 
denotes the number of hidden units, is passed to a fully connected 
layer which applies an affine transformation:
\begin{equation}
	\by = \bW \bh_T + \bb,
\end{equation}
where $\bW \in \mathbb{R}^{2 \times n_h}$ and $\bb \in 
\mathbb{R}^{2}$ are the weight matrix and bias vector of the fully 
connected layer, respectively, and $\by \in \mathbb{R}^2$ is the 
resulting vector of class scores (logits). The softmax 
layer then maps $\by$ to a probability distribution over the two 
model classes:
\begin{equation}
	p(i \mid \cZ_T) = \frac{e^{y_i}}{\sum_{j=1}^{2} e^{y_j}}, 
	\quad i \in \{1, 2\},
\end{equation}
and the predicted class label is assigned as:
\begin{equation}
	\hat{i} = \arg\max_{i \in \{1,2\}} p(i \mid \cZ_T).
\end{equation}

\section{Simulation Results} 
\label{sec:Results}
To quantify the performance trade-offs between the EM and LSTM 
classifiers, we conducted a series of controlled simulations using 
synthetically generated datasets. Specifically, a binary 
classification problem is considered, in which the classifier must 
determine which of two univariate state-space models generated a 
given observed sequence. The two models share identical parameters 
except for the noise parameters under study, making the 
classification task solely dependent on their statistical 
differences. The scalar SSM used in all experiments is:
\begin{equation} 
	\label{eq:scalar_ssm}
	\begin{aligned}
		x_{k} &= x_{k-1} + v_k, \\
		z_k   &= x_k + w_k,
	\end{aligned}
\end{equation}
where $v_k \sim \mathcal{N}(0,Q)$ and $w_k \sim \mathcal{N}(0,R)$, 
corresponding to $F=H=1$ in~\eqref{eq:gen_ssm}. Three classifiers 
are evaluated:
\begin{itemize}
	\item \textbf{True + Kalman LRT:} The true parameters and 
	underlying model structure are assumed known. The LRT is 
	applied directly using the true parameters. This classifier 
	is unrealizable in practice and serves as a reference for 
	the best achievable classification performance.
	
	\item \textbf{EM + Kalman LRT:} The model structure is assumed 
	known, but the true parameters are unknown. A separate set of 
	model parameters is estimated from $N_{\rm train}$ training 
	sequences of length $T$ for each class using the EM algorithm. 
	At test time, each of the $N_{\rm test}$ test sequences is 
	classified via the Kalman filter LRT using the estimated 
	parameters.
	
	\item \textbf{LSTM:} Both the true parameters and the model 
	structure are assumed unknown. An LSTM network is trained 
	directly on $N_{\rm train}$ labeled sequences of length $T$, 
	without any structural assumptions about the underlying model. 
	The trained network is used to classify $N_{\rm test}$ test 
	sequences.
\end{itemize}
Classification performance is measured by overall accuracy:
\begin{equation}
	\text{Accuracy} = \frac{\mathrm{TP} + \mathrm{TN}}
	{\mathrm{TP} + \mathrm{TN} + \mathrm{FP} + \mathrm{FN}},
\end{equation}
where $\mathrm{TP}$, $\mathrm{TN}$, $\mathrm{FP}$, and 
$\mathrm{FN}$ denote true positives, true negatives, false 
positives, and false negatives, respectively, with Model~2 
designated as the positive class. Since all datasets are strictly 
class-balanced, accuracy serves as an unbiased performance 
metric. All results are averaged over $N_{\rm MC} = 100$ Monte 
Carlo runs, with shaded bands indicating $\pm 1$ standard 
deviation.

\subsection*{EM Implementation}
The EM algorithm is implemented with 50 independent random 
restarts. For each restart, parameters are initialized as 
$F, H \sim \mathcal{U}(0.5, 1.5)$ and $Q, R$ are drawn 
log-uniformly from $[10^{-6}, 10^{-2}]$. A restart terminates 
upon reaching 50 iterations or when parameter updates fall below 
a tolerance of $10^{-7}$. The parameter set maximizing the 
training log-likelihood across all restarts is retained for 
classification.

\subsection*{LSTM Training Configuration}
The LSTM layer is configured with $n_h = 16$ hidden units. 
Training is performed using the Adam optimizer with an initial 
learning rate of $1 \times 10^{-3}$, a mini-batch size of 32, 
and a maximum of 100 epochs. A gradient clipping threshold of 
1 is applied to prevent exploding gradients. The network is 
retrained independently for each Monte Carlo run.

\subsection{Sensitivity to Task Difficulty}
\label{sec:exp_task_difficulty}
This experiment evaluates classification performance as a 
function of task difficulty, controlled by varying the noise 
parameters of one model while keeping the other fixed. Two 
cases are considered:
\begin{enumerate}
	\item \textbf{Process noise variation:} The measurement noise 
	is fixed at $R = 10^{-3}$, and $Q_1 = 10^{-5}$ while $Q_2$ 
	is varied over a logarithmic grid of 10 points from 
	$10^{-5}$ to $10^{-1}$.
	
	\item \textbf{Measurement noise variation:} The process noise 
	is fixed at $Q = 10^{-3}$, and $R_1 = 10^{-5}$ while $R_2$ 
	is varied over a logarithmic grid of 10 points from 
	$10^{-5}$ to $10^{-1}$.
\end{enumerate}
In both cases, when $Q_2 = Q_1$ (or $R_2 = R_1$), the two 
models are statistically identical and classification reduces 
to random guessing. As the parameters diverge, the models 
become increasingly separable and the task becomes easier.

\begin{figure}[h]
	\centering
	\includegraphics[width=.9\columnwidth]{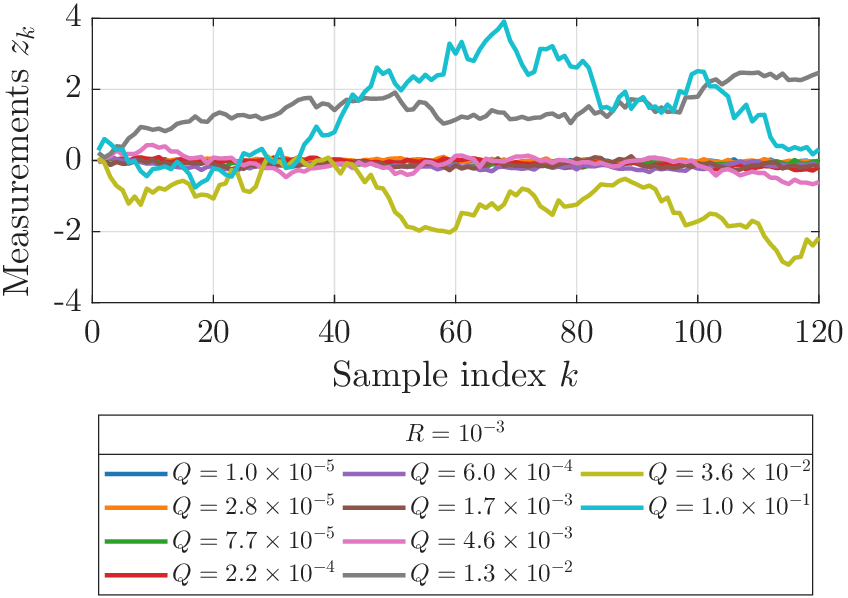}
	\caption{Sample observation sequences for varying process 
		noise variance $Q$, with fixed $R = 10^{-3}$.}
	\label{fig:data_Qratio}
\end{figure}

\begin{figure}[h]
	\centering
	\includegraphics[width=.9\columnwidth]{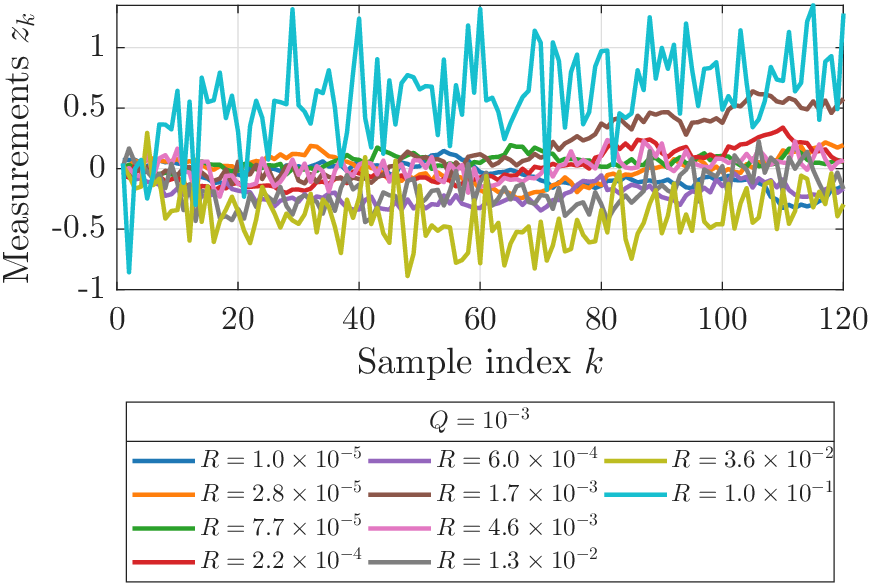}
	\caption{Sample observation sequences for varying measurement 
		noise variance $R$, with fixed $Q = 10^{-3}$.}
	\label{fig:data_Rratio}
\end{figure}

Sample realizations for selected parameter values are shown in 
Figs.~\ref{fig:data_Qratio} and~\ref{fig:data_Rratio}. 
Increasing $Q$ produces more dynamic state trajectories, making 
differences between sequences visually apparent 
(Fig.~\ref{fig:data_Qratio}). In contrast, varying $R$ has a 
less pronounced visual effect, as the measurement noise does 
not alter the underlying state dynamics, highlighting why 
classification based on $R$ alone is more challenging 
(Fig.~\ref{fig:data_Rratio}).


For each noise level, a balanced dataset of $N_{\rm train} = 500$ 
training sequences (250 per class) and $N_{\rm test} = 500$ test 
sequences was generated, each of length $T = 120$.

\begin{figure}[h]
	\centering
	\includegraphics[width=\columnwidth]{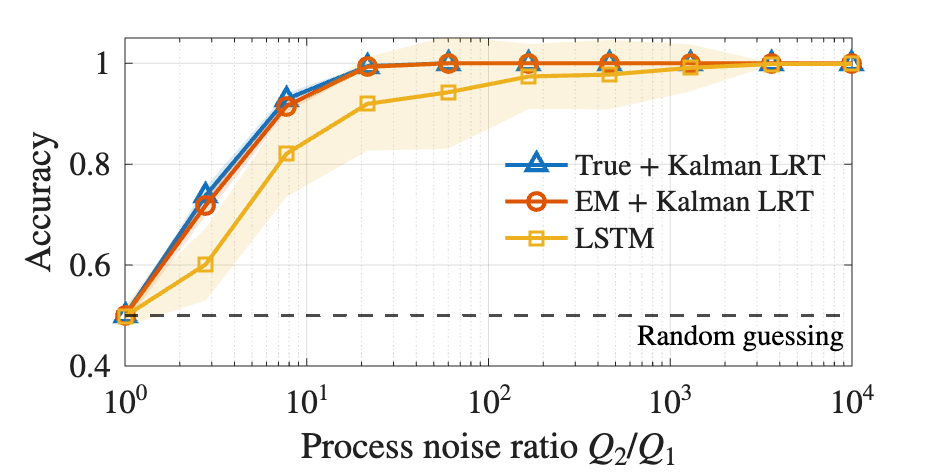}
	\includegraphics[width=\columnwidth]{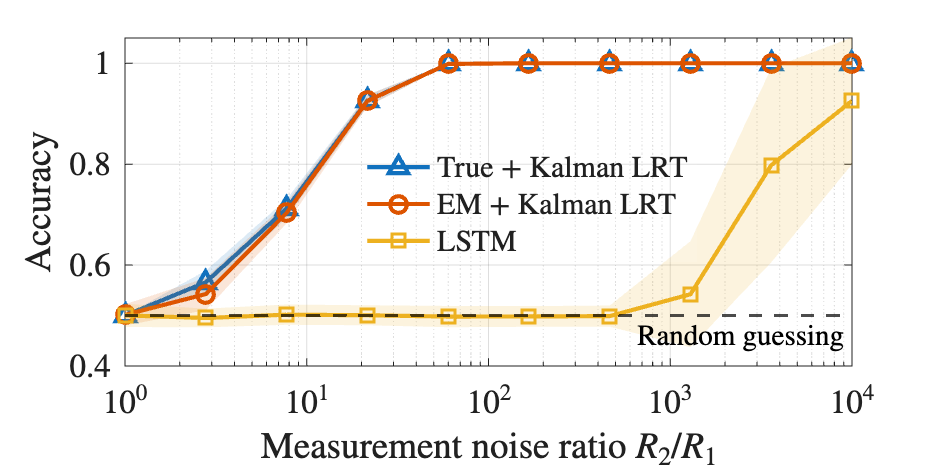}
	\caption{Classification accuracy as a function of noise 
		ratio. Top: process noise ratio $Q_2/Q_1$. Bottom: 
		measurement noise ratio $R_2/R_1$.}
	\label{fig:taskDifficulty}
\end{figure}

The mean classification accuracy and $\pm 1$ standard deviation 
across 100 Monte Carlo runs are shown in 
Fig.~\ref{fig:taskDifficulty}. When the ratio equals 1, the two 
models are statistically identical and both classifiers perform 
at chance level. As the ratio increases, both classifiers improve.

In the process noise experiment (top panel of 
Fig.~\ref{fig:taskDifficulty}), the EM classifier rapidly 
approaches the reference classifier as $Q_2/Q_1$ increases. 
This is expected, since the data are generated from linear 
Gaussian models and the EM algorithm is designed to estimate 
exactly this class of models. The LSTM classifier also improves 
with increasing ratio, though at a slower rate.

The measurement noise experiment (bottom panel of 
Fig.~\ref{fig:taskDifficulty}) proves significantly more 
challenging. While the EM classifier eventually converges to 
high accuracy at large ratios, the LSTM classifier struggles 
to learn the underlying differences and performs near chance 
level across much of the ratio range. This is consistent with 
the observation in Fig.~\ref{fig:data_Rratio} that varying $R$ 
has little visual effect on the observed sequences, making the 
task harder to learn from data alone. It should be noted that 
a relatively simple LSTM architecture was employed 
(Fig.~\ref{fig:lstm_arch}); incorporating convolutional layers 
or a deeper stacked LSTM may improve performance on this task.

\subsection{Impact of Sequence Length}
\label{sec:exp_sequence_length}
This experiment investigates how the accumulation of temporal 
evidence affects classification performance. The dataset size 
is fixed at $N_{\rm train} = 500$ training sequences (250 per 
class) and $N_{\rm test} = 500$ test sequences. The noise 
parameters are set to the moderately challenging regime 
($Q_1 = 10^{-5}$, $Q_2 = 2 \times 10^{-5}$, 
$R_1 = R_2 = 10^{-3}$), and the sequence length $T$ is swept 
over a logarithmically spaced grid. As noted in 
Section~\ref{sec:exp_task_difficulty}, this noise ratio 
constitutes a challenging task at $T = 120$.

\begin{figure}[h]
	\centering
	\includegraphics[width=\columnwidth]{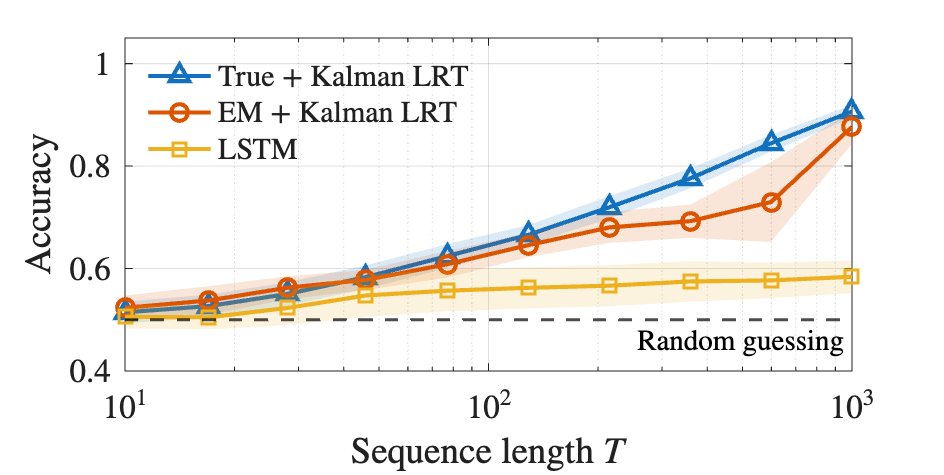}
	\caption{Classification accuracy as a function of sequence 
		length ($Q_1 = 10^{-5}$, $Q_2 = 2 \times 10^{-5}$, 
		$R_1 = R_2 = 10^{-3}$, $N_{\rm train}=500$).}
	\label{fig:accuracy_vs_Tsnap}
\end{figure}

The results are shown in Fig.~\ref{fig:accuracy_vs_Tsnap}. 
For short sequences ($T = 10$), both classifiers perform only 
slightly above the random guessing baseline. As $T$ increases, 
accuracy gradually improves as more temporal evidence 
accumulates. However, neither classifier reaches 100\% accuracy 
within the evaluated range. This is attributed to the high 
measurement noise ($R = 10^{-3}$), which is two orders of 
magnitude larger than the process noise and obscures the 
underlying state dynamics regardless of sequence length.

\subsection{Impact of Training Set Size}
\label{sec:exp_data_efficiency}
This experiment evaluates the sample efficiency of the LSTM 
classifier by investigating whether a larger training dataset 
allows it to overcome its lack of structural priors. The noise 
parameters are fixed at the same challenging regime as the 
previous experiment ($Q_1 = 10^{-5}$, $Q_2 = 2 \times 10^{-5}$, 
$R_1 = R_2 = 10^{-3}$), with sequence length $T = 120$. The 
total number of training sequences $N_{\rm train}$ is swept 
over a logarithmically spaced grid, with an equal number of 
samples per class.

\begin{figure}[h]
	\centering
	\includegraphics[width=\columnwidth]{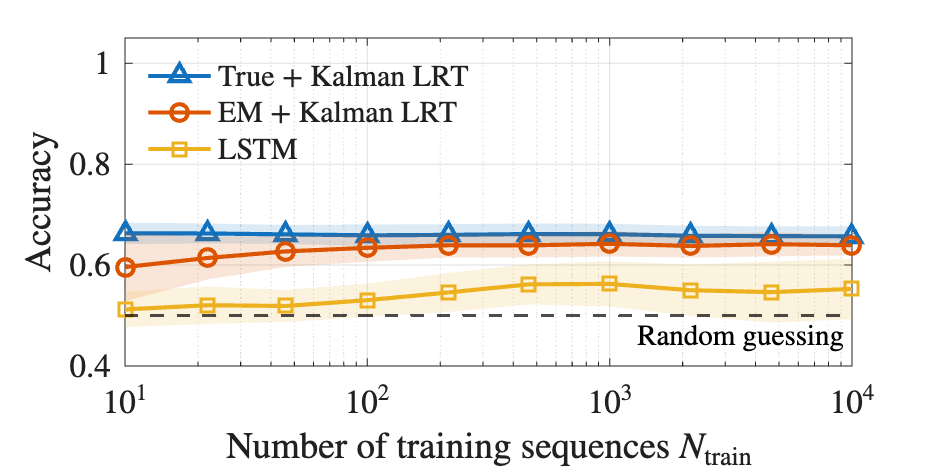}
	\caption{Classification accuracy as a function of the 
		number of training sequences ($Q_1 = 10^{-5}$, 
		$Q_2 = 2 \times 10^{-5}$, $R_1 = R_2 = 10^{-3}$, $T=120$).}
	\label{fig:accuracy_vs_Ntrain}
\end{figure}

The results are shown in Fig.~\ref{fig:accuracy_vs_Ntrain}. 
The reference classifier establishes an accuracy ceiling of 
approximately 65\%, reflecting the fundamental difficulty of 
the task under the given noise conditions. Increasing 
$N_{\rm train}$ does not allow the LSTM to overcome the severe 
measurement noise; its performance saturates well below the 
reference classifier and the EM classifier across the entire 
range of training set sizes. This demonstrates that the 
performance gap is not due to insufficient training data, but 
rather to the absence of structural priors that would allow 
the classifier to exploit the statistical properties of the 
underlying model.

\section{Conclusion}
\label{sec:conclusion}
This paper presented an objective performance evaluation 
framework for LSTM-based time-series classification, benchmarked 
against a model-based EM classifier. The evaluation was conducted 
on a binary classification problem involving two scalar linear 
Gaussian state-space models differing in their noise statistics, 
where the Kalman filter LRT with true parameters serves as a 
reference for the best achievable classification performance.

The results indicate that the EM classifier, which exploits the 
known model structure, performs strongly under the experimental 
conditions considered. However, the simulation conditions favour 
EM: the data are generated from the exact model class that EM 
assumes, and the model order is known. In practice, the true 
model structure is rarely known exactly, and model mismatch can 
degrade the performance of model-based approaches compared to 
data-driven alternatives~\cite{forti2023model}. Moreover, 
deriving exact analytical models for complex, nonlinear, or 
poorly understood systems is often intractable, which is 
precisely the regime where data-driven approaches are most 
valuable.

The LSTM classifier, operating without any model assumptions, 
requires a larger separation in noise statistics and more 
training data to approach the performance of the EM classifier. 
Since no structural assumptions are imposed, the network must 
learn the statistical structure of the latent dynamics from data 
alone, which is less efficient than exploiting known model 
structure. Nevertheless, once trained, an LSTM network performs 
classification via a single forward pass, avoiding the iterative 
parameter estimation, matrix inversions, and initialization 
sensitivity of EM, making it attractive in real-time and 
resource-constrained applications~\cite{liu2022tiny}.

The central challenge in modern time-series analysis is 
therefore not simply choosing between model-based and 
data-driven approaches, but recognising when to leverage 
established physical models and when the complexity of the 
environment renders such models inadequate. The proposed 
evaluation framework provides a principled basis for making 
this assessment, and can be extended to benchmark any deep 
learning architecture for time-series classification or 
forecasting against a theoretically grounded baseline. Future 
work will consider extending the framework to state and 
parameter estimation problems, multivariate and non-Gaussian 
noise models, and hybrid model-based deep learning 
architectures that combine the complementary strengths of both 
approaches~\cite{forti2023model}.

\bibliographystyle{ieeetr}
\bibliography{Literature}

\balance

\end{document}